\documentclass[runningheads]{llncs}
\usepackage[T1]{fontenc}
\usepackage{graphicx}
\usepackage{csquotes}

\begin{document}
\title{It does what it says on the tin: safe synthetic data from coarsened margins}
\titlerunning{It does what it says on the tin}

\author{Gillian M Raab}
\authorrunning{Gillian M  Raab}

\institute{University of Edinburgh and the Scottish Centre for Administrative Data Research }
\maketitle              
\begin{abstract}
This version will be made available on USB sticks to participants at PSD2026 at Cadiz, Spain, Sep. 30 - Oct. 2, 2026,  and presented at the meeting.  See https://crises-deim.urv.cat/psd2026/program.  It will not be published in the proceedings.

This paper proposes a method of creating synthetic data (SD) that will have two important advantages for the user. The first is transparency: the person in receipt of the SD will know which of the relationships between variables in the original data will be approximately maintained in the SD. The second is a guarantee that the SD is derived from information that has already been judged to be free of disclosure risk. This is achieved by first defining and calculating the margins where relationships between variables will be maintained in the SD. Each margin is then subject to statistical disclosure control (SDC) to the standards defined by the data custodian, e.g. top-coding and bottom-coding, combination of small categories and/or modifying small counts. Tables calculated from confidential data are often subject to SDC by removing details of all counts below a defined disclosure limit (e.g. counts below 10 are replaced by \textit{under10}).  In this methods all margins are  further adjusted, to prevent differencing attacks, by coarsening all counts in the table to multiples of the disclosure limit.  These adjusted margins are used to create SD by the Iterative Proportional Fitting (IPF) algorithm. The practical steps involved in creating this type of SD are illustrated using data from the 1901 Census of Scotland. 

\end{abstract}
\section{Introduction}
\subsection{Overview}
The creation of fully  synthetic data (FSD) was first proposed   as a privacy enhancing technology by Rubin \cite{Rubin} in 1993.  The term ``fully'' is here taken to mean that all records for all variables in the potentially disclosive ground truth (GT)\footnote{This acronym is used instead of original data (OD) to acknowledge that the OD is often pre-processed to correct obvious errors or inconsistencies and make it ready for analysis before the SD are created.} data are replaced by values generated from statistical models. It was hoped that the creation of synthetic versions of administrative data sets held by national statistics agencies (NSAs) might help to widen access to the information that can be obtained from such data and thus allow public sector data to be used to influence policy for the public good.

This paper considers the ways that SD can help this goal to be achieved, while assuring the public and data custodians in NSAs that releasing SD will not risk any breaches of the privacy of individuals whose records are held in the GT. We are concerned here with high-fidelity SD, as defined in \cite{lowfid}, that allows an analyst to reach the same conclusions from the SD as they would have obtained from the GT. The method we use is Iterative Proportional Fitting (IPF) \cite{DandS,Fienberg} that recreates the proportions in a cross-tabulation of categorical variables from a model defined by a series of known margins. This method was made available for the creation of SD as part of the \textit{simpop} package \cite{Templ} and is also now one of the methods available in the \textit{synthpop} package \cite{Nowok}. It has also been proposed as part of an auditable framework for releasing synthetic data \cite{framework}, and it has been  adapted to create SD from margins made to comply with differential privacy (DP) \cite{RaabDP,owen}. We adapt the method here to  create SD from margins that have been subject to SDC procedures.

\subsection{How can SD help to widen access?}
Data bases of administrative data are generally held in secure environments  where they can only be accessed by staff of the NSAs who analyse them to create national statistics and reports. Many agencies now have trusted research environments (TREs) where policy analysts, social scientists and others can access the data under strict supervision, without internet access and where research results can only be taken outside the safe setting after TRE staff have checked the output for disclosure risk \cite{Ritchie2007,frPSD2024,smithsdc}. Processes of obtaining access to TREs can be long and difficult. A TRE is a difficult work environment as it inhibits joint working for the interpretation of data. There are a variety of possible ways that SD can be used to widen access to data in the TRE.
\begin{enumerate}
\item{\textbf{REPLACEMENT} Use of the FSD instead of the GT to draw policy conclusions, possibly with the option of having results confirmed by running them on the GT before results are published.}
\item{\textbf{PLANNING} To use the SD  to understand the variables in the data and formulate research plans.}
\item{\textbf{TEACHING} FSD can be used to create data sets for teaching researchers about the data held in the TRE and how to use it.}
\item{\textbf{CODE WRITING} Giving potential users of TREs access to a file with the same structure (variable names, types and details) as the original data, to allow code to be developed that they can use when access to the GT is permitted.}
\end{enumerate}

An additional aspect is whether the SD can be accessed outside the TRE, and if so under what conditions: freely available, only to identified individuals or under an end-user license agreement (EULA) where the recipient of the FSD signs up to conditions of use.  The UK Data Service defines these three levels of access as \textbf{open} data, \textbf{safeguarded} data and \textbf{controlled} data\footnote{https://ukdataservice.ac.uk/find-data/access-conditions/, accessed 26/5/26}.  Data custodians decide what level of access is appropriate for their SD,  based on their perception of its potential to lead to a data breach or to a loss of reputation to the agency.

\subsection{Does FSD pose a disclosure risk to individuals?}
Initially it was thought that, since no records in FSD could be related to an identifiable  individual, FSD was not personal data, as defined in the General Data Protection Regulation (GDPR), Article 4(1) and thus would not pose any disclosure risk to the individuals in the GT\footnote{https://gdpr-info.e,. accessed 26/5/2026}. However, this view has changed for SD that is created using data from real individuals to create the models that are used to generate the SD. The UK Information Commissioner's guidance on SD now states:
\begin{displayquote}
...you will generally need to process some real data in order to determine realistic parameters for the synthetic data. Where that real data can be related to identified or identifiable individuals, then the processing of such data must comply with data protection laws.\footnote{https://ico.org.uk/for-organisations/uk-gdpr-guidance-and-resources/artificial-
intelligence/guidance-on-ai-and-data-protection/how-should-we-assess-security-and-
data-minimisation-in-ai, accessed 24/05/2026}    
\end{displayquote}

When no personal data has been used as input to the synthetic data generator, GDPR does not apply \cite{DARE2}. An example is low-fidelity SD, created only from publicly available meta-data, but this method produces very low-fidelity SD that does not attempt to approximate even univariate distributions, and can only be used for code development.
 
 In the last decade there has been an explosion of developments in methods for creating FSD, many based on machine learning (ML) methods. A large number of websites now offer tools to create FSD. These can be either free open-source methods or commercial sites. Some of each type claim that their methods can produce data that will reproduce the results that would be obtained from the GT without any disclosure risk. For example, the open-source library Synthetic Data Vault claims that ``Artificial data give the same results as real data — without compromising privacy''\cite{SDV}.
 
 Unsurprisingly, NSAs take a more cautious view of the ability of FSD to give the same results as real data and to give protection against any disclosure of personal data they hold. There have been few releases of FSD from NSAs outside a TRE as \textbf{open} or even \textbf{safeguarded} data, except for those intended only to be used for code development. SD created by the  US Government \cite{synLBD,synSIPP} are only released within a TRE\footnote{https://www.census.gov/programs\-5surveys/sipp/guidance/sipp\-5synthetic\-beta\-data\-product.html, accessed 26/5/2026, details the restrictions on their use}. 
 Statistics New Zealand, who were early pioneers of methods of creating FSD \cite{NZ} now appear to have only one project that provides open FSD for 6 variables for a now-discontinued income survey. In the UK there appear to be only a few examples of the release of FSD,  accessible outside a TRE, that could be used for any purpose other than code development.  Two of these are the simulacrum data on cancer outcomes available from Public Health England\footnote{See https://simulacrum.healthdatainsight.org.uk/, accessed 26/5/2026} and Primary Care Data from the Medicines and Healthcare products Regulatory Agency\footnote{See https://www.cprd.com/data/synthetic-data, accessed 26/5/2026}.
 In both cases the web sites from which they can be accessed carry extensive warnings about the possibility that their results may differ from what would be found in the GT data.
 Also, users of the Scottish Longitudinal Study can request synthetic extracts after agreeing to an EULA \footnote{https://www.lscs.ac.uk/}, as can users of education data provided by DUO in the Netherlands 
 \footnote{See https://duo.nl/open\_onderwijsdF25 minutesata/synthetische\-data.jsp , accessed 28/05/2026}.  Drechsler and Haensch \cite{DandH} provide a comprehensive evaluation of the practical applications of SD in the 30 years since it was first proposed.
 \subsection{UK policy on the release of synthetic versions of public sector data}

In the UK several publicly funded bodies (ADR UK, HDR UK, DARE UK, SCADR) \footnote{https://www.adruk.org/ https://www.hdruk.ac.uk/ https://dareuk.org.uk/ https://SCADR.ac.uk} have been established to facilitate researchers' access to public sector data. ADR UK's current recommendation on the role of FSD \cite{ADRUK}, is that data providers make low-fidelity SD available to researchers. This is defined as FSD that does not attempt to reproduce the relationships between variables in the GT \cite{lowfid}. It can be produced in two different ways. The lowest fidelity is obtained by using only publicly available meta-data on the codes and ranges of variables in the data. Alternatively, the univariate distributions of the variables can be used to produce SD that aims to preserve the univariate distributions. In this second case the univariate distributions can first be checked to ensure a low risk of disclosing information about any individual in the GT.  In a recent consultation report \cite{DARE} data owners favoured this approach. Two examples of this  type of low-fidelity FSD are now available from the Longitudinal Eduction Outcomes data set (LEO)\footnote{See https://datacatalogue.ukdataservice.ac.uk/studies/study/9505, accessed 26/5/2026} and from Hospital Episode Statistics provided by NHS England, as part of their artificial data pilot\footnote{see https://digital.nhs.uk/data-and-information/data-tools-and-services/data-services/hospital-episode-statistics, accessed 16/5/2026}.
 
 \section{Methods of SD creation with limited disclosure risk}
 \subsection{By modifying methods until calculated disclosure risk is low}
A possible approach to allowing the release of FSD, while mitigating any privacy risk, would be to assess its disclosure risk and adjust the data and the methods by which it is produced until its disclosure risk is acceptable according to some specified measures of disclosure risk. This is analogous to proposed methods of ensuring the utility of SD \cite{Raab2021}. The disclosure risk of the SD can be modified by SDC procedures applied to either the GT data or to the SD before its release, and by modifying the method used to create it.

This approach was suggested in \cite{RaabPSD2024}; however, the privacy metrics proposed in \cite{RaabPSD2024} are very limited. They are measures that assess whether a naive intruder who believes the FSD to be the GT can 
gain correct information about a person in the GT with certain known characteristics, often referred to as ``keys'' or ``quasi-identifiers''.  They do not address the risk posed by a potential intruder with the competency to use advanced methods to interrogate the FSD.  Given the increasing availability of AI methods that can scrape information from the internet, it is understandable that UK custodians of public sector data do not wish to attempt to go down this road. Instead an easier choice is the conservative approach to SD, as discussed for the UK above.

\subsection{Creating SD from non-disclosive summary statistics}
The low-fidelity SD created from univariate distributions is one example of this approach. It can be used for developing code, but its value for research planning or for creating teaching data sets is limited.  Another approach, designed for creating teaching data sets, is to use the results of published analyses of GT data, along with univariate margins, to generate FSD with genetic algorithms. It is described by \cite{ElliotPSD2024} and has been used to create a teaching dataset from the ONS Annual Survey of Hours and Earnings, linked to 2011 Census data. \cite{ashe_census2011}.

Methods of creating SD can be made to comply with differential privacy (DP):  a formal guarantee that limits the influence of a single record on each of the summary statistics used in the creation of the SD. 
The release of outputs from the the USA 2020 Census  have been subjected to a DP audit \cite{Abowd2018} whose practical consequences have 
been criticised in \cite{Rug,Wink} and by many others. Drechsler's review of the use of DP by government agencies \cite{Dre} raises many challenges that need to be addressed in considering using DP compliance as a criterion for releasing data to the public. In section 3.1 we discuss how  methods based on DP margins could be compared to the method proposed in this paper.

In 2018 the National Institutes for Standards and Technology (NIST) ran a challenge that invited teams to create SD that complied with DP for a specified set of DP parameters. The winning team used minimum spanning tree (MST) methodology to identify a set of margins that, along with the univariate distributions of all variables, will capture all the relationships between variables in the GT data \cite{McKenna}.  The selected margins are then made DP by the addition of a controlled amount of noise determined by the DP parameters $\epsilon$ and $\delta$. The synthetic data are then generated from the distribution of the MST defined by these altered marginals. This approach has also been used by the UK Office of National Statistics to create synthetic data from Census data linked to deaths \cite{ONS}.

SD can be created using  IPF, from margins with noise added to make them DP \cite{RaabPSD2024}.  This method was also mentioned by \cite{Daniel} as part of the SASPSA: Select, Assess, Privatise, Synthesise, Audit procedures that allows stakeholders to take part in decisions about which relationships should be maintained in the SD.	

The method we propose below has features in common with each of these methods. Like the method used to create teaching data sets \cite{ElliotPSD2024} it uses outputs that are accepted as non-disclosive by SDC procedures.
It resembles the DP methods \cite{McKenna,ONS} by MST or by IPF \cite{RaabPSD2024,Daniel}, in that it uses a set of altered margins to define the distribution.

\section{Synthetic data from published coarsened margins.}
\subsection{The method}
The method consists of creating SD by IPF from a set of margins calculated from the data and altered so as to satisfy SDC criteria for disclosure control.
The standard method of SDC for tables consists of identifying table entries below a disclosure limit (typically 5 or 10). When multiple tables are released 
this may not be sufficient because cells below the disclosure limit may be derived by differencing the tables or deriving cell values from known totals or margins. Procedures to avoid such disclosures are described by the team at the University of the West of England \cite{Ritchie2007,frPSD2024,smithsdc}. They suggest that disclosure control staff should examine all possible combinations of published tables to check them for disclosure by differencing. If a large set of tables are to be used this is  a very tedious task  

A more practical solution for the altering the margins of the tables to use as the basis for IPF  would be to coarsen the data by expressing all the counts in the margins as multiples of the disclosure limit. For a disclosure limit of 10, all the counts in the margins could be replaced as follows: counts 0-9 to 10, counts 10-19 to 20, counts 20-29 to 30... etc.  Differencing such tables would not correspond to the differences in the original tables.

The proposed method consists of the following steps. We describe the creation of three sets of altered margins: \textbf{disclosure controlled} individual tables, \textbf{coarsened} tables and \textbf{adjusted coarsened} tables.
\begin{enumerate}
\item{Carry out SDC procedures on the GT data, such as top- and bottom-coding for numeric variables and  group them into ranges, form merged categories of any level of a variable with a small count\footnote{these steps are often  done for data that will be available in the TRE}.}
    \item{Decide on the margins to be preserved in the SD and create them from the GT data.}
    \item{Check the margins for small cells below a 
    disclosure limit specified by the TRE, often the value of 5 or 10, as suggested for SDC checking \cite{Ritchie2007,frPSD2024,smithsdc}.}
    \item{Adjust the margins by replacing counts in the tables that are below the disclosure limit with the disclosure limit. These are the \textbf{disclosure controlled margins}.}
    \item{Each such margin taken by itself, would be considered non-disclosive in SDC checks. But if the total sample size is known or if more than one margin is published, then cells below the disclosure limit can be identified by differencing, and margins need to be  be adjusted to prevent disclosure from multiple tables.
    Rather than seeking out disclosive margin combinations we suggest coarsening all the counts in the margins by replacing them with multiples of the disclosure limit. These are \textbf{coarsened margins}.}
    \item{Each \textbf{coarsened margin} will generally sum to more than the sample size of the GT data. A small count can then be subtracted from all the cells in every margin to make the mean table total close to the number of records in the GT. These are \textbf{adjusted coarsened margins}.}
    \item{Obtain a fit to the joint distribution of all the variables by IPF for each of the three sets of margins.} 
    \item{For each of these three,  generate SD as samples of any desired size from a multinomial distributions with probabilities given by the fitted IPF proportions.}
\end{enumerate}

When any of the margins have been altered, the margins will no longer be compatible with each other. This is also the case for using IPF on DP margins. The case of incompatible margins has been discussed, usually in the context of proportions derived from samples from a population where some marginal totals of the population are known \cite{Stephan}. This has been extended \cite{LittleWu} to the case of systematic differences between the sample and the population. In both cases the IPF estimates are shown to converge and to be consistent.

 Our case is different because the differences between the altered counts and those from the GT data have a distribution that depends on the counts in the tables from the GT data,  This has not prevented the IPF algorithm from finding an acceptable fit to the  margins for any of the examples we have tried, though further investigation of when problems might occur should be carried out.

The decision on which margins to fit can be taken in various ways. An analysis of the GT with log-linear models might be carried out to decide on important relationships between variables. Another approach, that we illustrate below, is to start with fitting all two-way margins and then explore whether there are any three-way margins that have not been fitted adequately.

Where any of the margins contain zero counts, a decision needs to be made as to whether these are structural zeros: a case that could never arise, such as answering ``yes'' to ``do you have a driving license?'' for people under 18 in the UK. Non-structural zeros should be counted as disclosive and suppressed by replacing them with the disclosure limit. Similarly, some missing values are determined by the structure of the data; for example, number of years smoked must be missing for non-smokers. At stage 5 of the process described above any structural zeros in the margins should be replaced  by zeros.

Should the NSA publish the sets of margins that have been used to create the SD sets? This would be recommended to increase the transparency of the method, but it could mean publishing a large number of tables. Further work is required to consider how such publication might affect disclosure  risk.

\subsection{Limitations}
The two limitations of the IPF algorithm are the need to create a large cross-tabulation of all the variables, and the time for the large number of iterations around many tables that can sometimes be needed for the algorithm to reach convergence.  The eleven variables in the example discussed below generated a table in R with just over 50 million cells, analysed on a modest Windows laptop \footnote{Intel(R) Core(TM) i5-6300HQ CPU @ 2.30GHz (2.30 GHz) processor with 12Gbytes RAM}, but adding more variables  would go beyond the available memory.  With better computational resources, and/or different software, this method could be applied to data sets with more variables. The tolerance for the maximum difference between the proportions to define convergence can be adjusted to speed up the fit and larger values still seem to give acceptable results.

When IPF is carried out with higher-order margins, it is important to include the univariate margins for any variables that do not appear in any of other margins. Failure to include them will produce univariate distributions from the SD with equal proportions in each category. Another possibility would be to include one-way margins along with higher-order margins even when they do appear in the higher-order margins. This was attempted with the example below, but did not improve the utility and may increase disclosure risk.

\section{A worked example}
\subsection{Data from the 1901 Census of Scotland}
Data on heads of household in the 1901 Census, that would have been disclosive at that time even if anonymised, are used to illustrate the method. It consisted of 51,074 records for heads of household in the City of  Edinburgh. Eleven variables from this data set were selected to use to create the SD. The numeric variables were categorised as shown in Table 1.  Initially, all one-way and two-way tables were screened for 
counts below the disclosure limit, taken as 10. Looking at one-way tables revealed 7 cases where the number of family members over 15 was zero. On checking back to the original records, these proved to be families where the children had been left under the care of servants. These were removed from the file, leaving 51,067 records.  A further 3 records were removed for the small occupation group ``fishers'' leaving 51,064 records\footnote{In 1901 fishers would have resided in the port of Leith, rather than the City of Edinburgh.}. 

\begin{table}[h]
\caption{Details of 11 variables selected for synthesis after data curation, as described below. The categories for ``occlab1'' ranged from the largest two groups ``building and construction'' (6085 12\%)  and ``food, tobacco, drink and lodging'' (10\%) to the smallest groups such as ``defence of the country'' (194) and ``brick, cement, pottery and glass'' (251).}
\centering
\begin{tabular}{|r|c|r|r|}
  \hline
variable & number of levels & labels \\ 
  \hline
sex & 2 & F M \\ 
  age & 8 & under20 20\_29 30\_39 40\_49 50\_59 60\_69 70\_79 80+ \\ 
  mar\_stat & 5 & ``Married'' ``MarriedSpouseAbsent'' ``NotKnown'' ``Single'' ``Widowed'' \\ 
  occlab1 & 23 &  \\ 
  employ & 3 & BLANK Employer Worker \\ 
  inactive & 6 & Other OwnMeans Retired Unemployed Working xmiss \\ 
  ctry\_bth & 3 & ENG OTH SCT \\ 
  nservants & 2 & none any \\ 
  nfamgteq15 & 6 & 1 2 3 4 5 6+ \\ 
  nfamlt15 & 7 & 0 1 2 3 4 5 6+ \\ 
  totrooms & 6 & 1 2 3 4 5+ xmiss \\  
   \hline
\end{tabular}
\end{table}
The following steps were taken to create the GT from the original records. Variables with small categories were regrouped. Country of birth, with was simplified by grouping those not born in England, Scotland or Ireland to ``OTH''. For the variable ``inactive'', the 9 ``paupers'' were combined with the unemployed and the categories ``pension'' and ``formerly occ'' were combined with ``retired''. 
Numbers of family members and total rooms in the household  were top-coded, as shown in Table 1. These are tasks that would normally be part of making data research-ready \cite{res_ready}.  The data on age was top- and bottom-coded with limits at 18 and 88\footnote{As these data were for heads of households there were very small numbers under 18.}. The resulting ages were then grouped, as shown in Table 1. Finally, the data on the number of servants in the household was simplified to whether there were any servants at all. This was done because one of the use-cases envisaged was an analysis of which households employed servants, and for the purpose of this paper a simplified model would be easier to explain.

Missing values were included as a separate category where they occurred, shown as ``xmiss''. No further efforts were made to correct the data and remove any inconsistencies. The expected use-case for this type of SD is for research planning or teaching. People using the SD would be expected to make their own decisions about what to do about missing values and whether to remove inconsistencies.

\subsection{Creating and adjusting margins}
To illustrate the method, it is assumed that tables created from this data would be subject to output checking to ensure that no counts in any table used to create the SD were below a limit of 10. After the data curation described above, no one-way table had a count below the limit, but 29 of the 55 two-way tables had one or more counts below the limit. Of the 1,528 cells in these 29 tables 243 (16\%) were below 10.  New sets of all two-way tables were created from these margins as follows:

\begin{itemize}
	\item{\textbf{Disclosure controlled} Any counts below 10 in the table replaced with 10}
	\item{\textbf{Coarsened} to multiples of 10 ,  0-9 to 10, 10-19 to 20,  etc.}
	\item{\textbf{Adjusted Coarsened} Adjust margins by subtracting a number from the counts in each cell of every table to make the average counts per margin close to the original records}.
\end{itemize}

The average count in the disclosure-controlled margins was just 28 larger than the sample size of the GT data. However, the coarsened margins had average totals that were 220 greater than the GT size. By subtracting 5 from the cells of all tables this difference was reduced to 27.  Table 2 illustrates one two-way margin with 3 counts below 10 and how it would appear in each of the three sets of altered margins.

\begin{table}[h]
\centering
\caption{Cross tabulation of two employment variables for 51064 heads of households in 1901 Census data for the City of Edinburgh and modified as we described here.}
\begin{tabular}{|r|r|r|r|r|r|r|r|}
  
  \multicolumn{8}{c}{Original GT data} \\
  \hline
 & other & own means & retired & unemployed & working & xmiss & \textbf{Totals} \\ 
  \hline
BLANK & 639 & 2605 & 1489 & 35 & 8519 & 4436 & 17723 \\ 
  Employer & 10 & 1 & 12 & 0 & 3398 & 3  & 3424\\
  Worker & 35 & 29 & 71 & 42 & 29709 & 31 &29917\\ 
  \hline
  \textbf{Totals} & 684 & 2635 & 1572 & 77 & 41626 & 4477 & 51064\\ 
  \hline
   \multicolumn{8}{c}{ }  \\   
   \multicolumn{8}{c}{Disclosure controlled with cells under 10 replaced by 10} \\
     \hline
   & other & own means & retired & unemployed & working & xmiss & \textbf{Totals} \\ 
  \hline
  BLANK  & 639 & 2605 & 1489 & 35 & 8519 & 4436 &1723 \\ 
  Employer  & 10 & 10 & 12 & 10 & 3398 & 10 & 3450 \\ 
  Worker  & 35 & 29 & 71 & 42 & 29709 & 31 & 29917 \\
  \hline
 \textbf{Totals}  & 684 & 2644 & 1572 & 87 & 41626 & 4477 & 51090 \\
 \hline
     \multicolumn{8}{c}{ }\\
    \multicolumn{8}{c}{Coarsened to multiples of 10, e.g. 0-9 to 10 , 10-19 to 20  etc.} \\
  \hline
   & other & own means & retired & unemployed & working & xmiss & \textbf{Totals}\\ 
  \hline
  BLANK & 640 & 2610 & 1490 & 40 & 8520 & 4440 & 17740 \\ 
  Employer & 20 & 10 & 20 & 10 & 3400 & 10 & 3470\\ 
  Worker & 40 & 30 & 80 & 50 & 29710 & 40 &29950 \\ 
    \hline
  \textbf{Totals}  & 700 & 2650 & 1590 & 100 & 41630 & 4490   & 51160  \\
  \hline
  \hline
   \multicolumn{8}{c}{ }\\
      \multicolumn{8}{c}{Coarsened data minus 5} \\
  \hline
   & other & own means & retired & unemployed & working & xmiss \\ 
  \hline
  BLANK & 635 & 2605 & 1485 & 35 & 8515 & 4435 & 17740 \\ 
  Employer & 15 & 5 & 15 & 5 & 3395 & 5  & 3440\\ 
  Worker & 35 & 25 & 75 & 45 & 29705 & 35& 29920 \\ 
   \hline
     \textbf{Totals}  & 685 & 2635 & 1575 & 85 & 41615 & 4475   & 51070  \\
   \hline
\end{tabular}
\end{table}

An SD set was generated from the fit to each of these three sets of margins using routines in the $mipfp$ package for R \cite{mipfp}. Each fit followed by the creation of the SD took around 25 minutes on the Windows 11 laptop described above.

Note that synthesis from the original data or from disclosure-controlled margins would risk disclosure of small cells if all the margins were published. By comparing it to the GT data we will be assessing whether the IPF model of two-way margins is an appropriate generative model for this data set. The other two SD sets should prevent any disclosure of small counts, even if all the margins are made available either publicly or to the recipients of the SD. 

\subsection{Comparing the utility of the three synthetic data sets}
To compare each of the SD sets with the GT, we start by evaluating how well they reproduce tables. The measure used is the standardised propensity score mean square error $SpMSE$ \cite{snoke}. The propensity score is calculated by combining the rows of the observed and SD and assigning an indicator variable with the value $1$ for synthetic rows and zero for original rows. The propensity score is then calculated from the predicted value of the indicator variable from the combined data, as the mean value of the squared difference of the predicted values from 0.5 for the case when SD and the GT have the same number of records, $N$\footnote{when numbers are unequal more complicated formulae are needed}.

For a table with $n$ cells containing GT counts of $y_i$ and synthetic counts $s_i$, $i = 1,...n$, the predicted value from the combined data is just $s_i/(s_i + y_i)$ for all the observations in that cell. Hence we get
\begin{equation}
pMSE = \frac{1}{N}\sum_{i=1}^k{(s_i + y_i){(s_i/(s_i + y_i) - 0.5)^2}}
\end{equation}
reducing to
\begin{equation}
pMSE = \frac{1}{8N}\sum_{i=1}^k{\frac{(s_i - y_i)^2,}{(s_i + y_i)/2}}.
\end{equation}
We can see that equation 2 has a form similar to Pearson's  $\chi^2$ statistic, but with the denominator replaced by the average of the observed and the expected. It was proposed as a measure for assessing SD by Voas and Williamson in 2001 \cite{VandW}. The $SpMSE$ is identical to Voas and Williamson's modified $\chi^2$ statistic divided by its degrees of freedom. If the model used to generate the SD is correct, in that the GT is a sample from the generative distribution implied by the synthesis model\footnote{Here a log-linear model with all two way interactions}, then the $SpMSE$ will have an expected value of 1.0. Experience with using this measure on SD has shown that no major differences can be identified between conclusions from the SD or the GT until $SpMSE$ values exceed 10 \cite{Raab2021}. 

Table 3 gives the $SpMSE$ for one-way tables  for the three SD sets. We can see that they are all well below 10, and to confirm the absence of any important differences, Table 4 compares the proportions in the GT and SD sets for the variable with the worst $SpMSE$. Note also that synthesis from coarsened, adjusted marginals gives the lowest $SpMSE$ values.

\begin{table}[h]
\caption{One\-way utility measures $SpMSE$ for 11 variables from SD created from 3 sets of two-way marginals}
\centering
\begin{tabular}{|c|ccc|}
  \hline
 & Disclosure &  & Coarsened  \\ 
Variable &  Controlled & Coarsened & adjusted \\ 
  \hline
sex & 0.57 & 1.68 & 0.22 \\ 
  age & 1.65 & 5.59 & 1.28 \\ 
  mar\_stat & 1.24 & 5.81 & 2.00 \\ 
  occlab1 & 2.78 & 2.36 & 1.53 \\ 
  employ & 2.04 & 0.34 & 0.38 \\ 
  inactive & 2.15 & 1.91 & 1.07 \\ 
  ctry\_bth & 1.52 & 2.38 & 1.28 \\ 
  nservants & 0.05 & 0.21 & 0.05 \\ 
  nfamgteq15 & 0.25 & 1.51 & 0.62 \\ 
  nfamlt15 & 1.49 & 1.60 & 1.56 \\ 
  totrooms & 0.42 & 2.31 & 0.98 \\ 
   \hline
\end{tabular}
\end{table}

\begin{table}[h]
\caption{Tables of proportions for variable mar\_stat for the original GT data and from SD created from two-way tables altered in different ways.}
\centering
\begin{tabular}{|r|rrrrr|}
  \hline
 & Married & Married spouse absent & Not known & Single & Widowed \\ 
  \hline
Original GT & 64.40 & 5.30 & 0.20 & 9.80 & 20.30 \\ 
  Disclosure Controlled & 64.20 & 5.30 & 0.20 & 9.90 & 20.40 \\ 
  Coarsened & 64.60 & 5.30 & 0.20 & 9.70 & 20.10 \\ 
  Coarsened adjusted & 64.00 & 5.50 & 0.20 & 9.80 & 20.50 \\ 
   \hline
\end{tabular}
\end{table}

The next step is to calculate the $SpMSE$ metrics for all 55 two-way tables.  Figure 2 illustrates the values for each SD set. The shading is scaled up to 10, with all values coming in just below this. The maximum value was 9.1 for one of the disclosure controlled tables. The two coarsened SD sets gave similar results with the mean $SpMSE$ being lower for the adjusted margins (1.3) compared to unadjusted (1.6).

\begin{figure}[h]
  \centering
  \includegraphics[width = 1\linewidth]{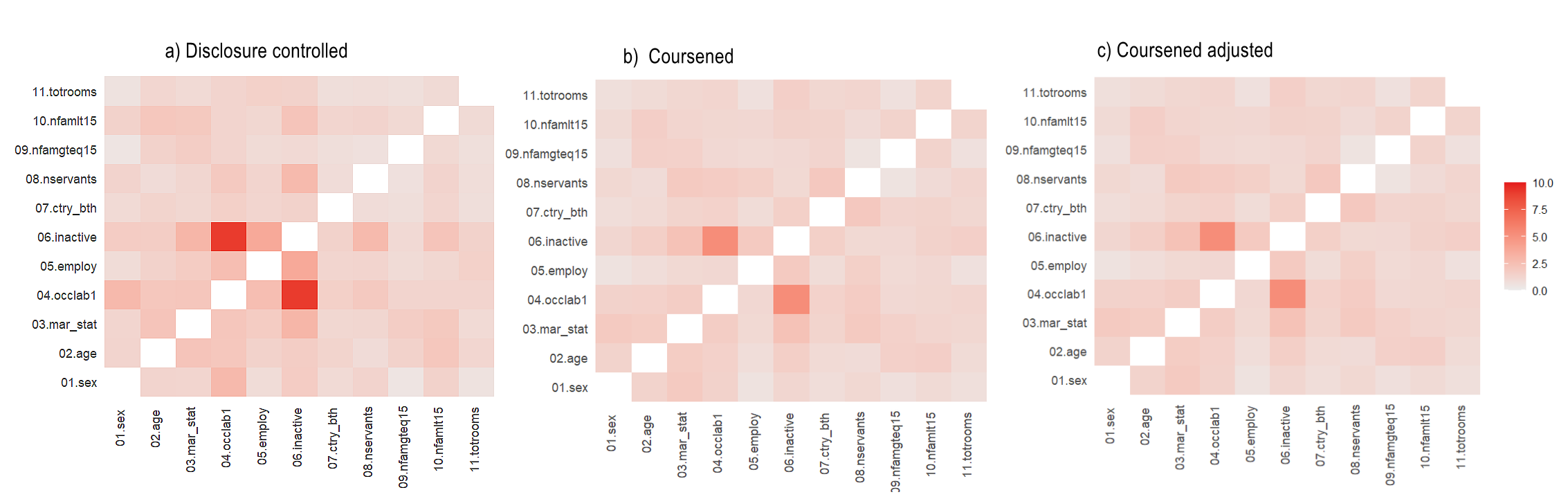}
  \caption{Plots of $SpMSE$ values for all two-way tables for each synthesis method.}
  \label{fig:f1}
\end{figure}

The results so far suggest that the SD based on coarsened and adjusted margins performs the best of these three methods, although they are all rather similar.  It is also the method that will allow the data for the margins used in the fitting to be released with minimal disclosure risk. Further results will be given for just this method alone.

The agreement between the GT and the SD was also assessed for three-way tables. No table gave an $SpMSE$ value over 10, but a few tables had values between 5 and 10, most of which involved sex and age. Examining female heads of households in the GT revealed a different age structure by sex and marital status. To examine this for the SD,  synthetic ages would need to be created from the grouped values in the SD. This could  easily be done by sampling from the ranges of the groups. If such analyses were of interest to those using the SD some three-way margins could be included in the IPF fit used to create the SD. 

A final check on the SD is to use it to investigate how it might be used to fit a statistical model to predict an outcome. The outcome selected was whether a household employed any servants, predicted by logistic regression. Such modelling usually proceeds by first examining the univariate between candidate variables and the outcome (any servants). As expected from the results above, the SD gave exactly the same picture as the GT. Then a series of models could be fitted to find one with the best explanatory power and the most helpful interpretation. None of the models explored showed a different picture for the SD compared to the GT.

A comparison between the fit for the SD and the GT is shown in Figure 2 for a model including sex,  age group, marital status and numbers of rooms in the house. Houses with 5+ rooms (baseline was 1 room) were more likely to have servants, as were single people and widows. Households headed by men were more likely to have servants than those headed by women. The mean confidence interval overlap, the ratio of the overlap of the 95\% confidence intervals to the average of their lengths,  was 68\%.
\begin{figure}[h]
  \centering
  \includegraphics[width = 1\linewidth]{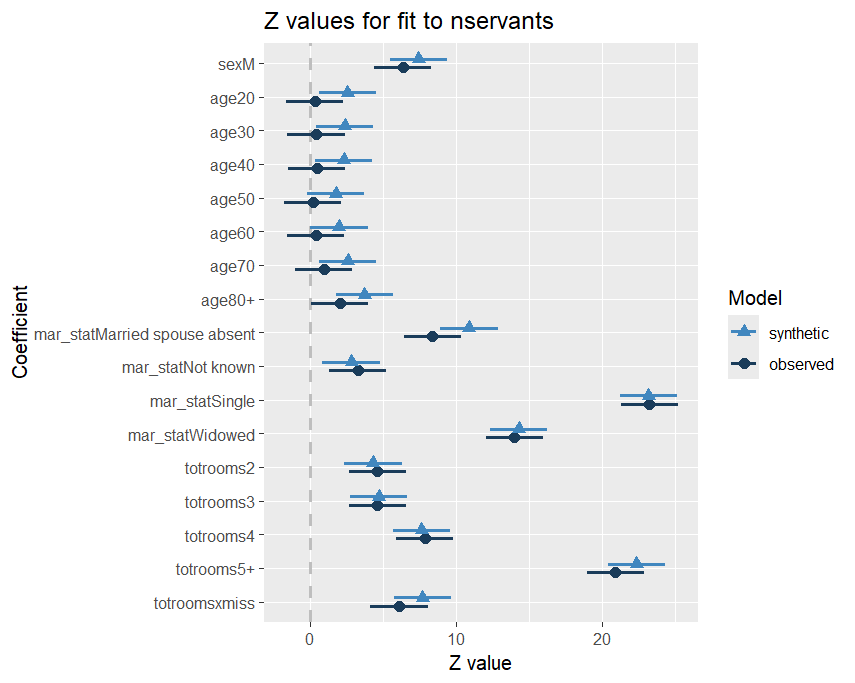}
  \caption{Comparison of Z-scores in a logisitic regression to predict which households have servants using the SD and the GT}
  \label{fig:f1}
\end{figure}

\section{Practical recommendations and further work}
The proposed method of creating SD from coarsened margins seems to have worked well for this example. But this paper is just a proof-of-concept excercise and much more investigation will be needed to explore it further. 
It has much in common with the method developed by the Manchester group \cite{ashe_census2011,ElliotPSD2024}. that now has an open source, easy-to-use, version available \footnote{EDASIDAGUI see https://synthdig.github.io/ accessed 10/5/2026}. Their method was developed for use in training courses, and it has been used for this. Training courses would be an obvious use for the methods proposed here. But an additional use would be data exploration.  If a machine were available with a large RAM, then data sets with more 
variables could be created. As is described here, the GT data needs first to be made research-ready before it is synthesised.

The method described in section 3 could have the potential to make SD available safely. The effort to program this method are minimal and its properties are transparent. Open-source code will soon be made available, but unanswered  questions remain.   While the disclosure control of the margins suggest that the SD would be free from disclosure risk, this should be checked by calculating disclosure risk metrics. The same procedure could be used for the same margins made to comply with DP.
A comparison of the utility of DP SD calculated from margins could also be investigated.  Methods that use DP margins attempt to identify a small number of margins to capture the joint distribution. This is because the total $\epsilon$ for the SD needs to be subdivided and allocated across all the margins. Dividing an  $\epsilon$ of 1 or even 10 equally between the 55 margins in our example added so much noise that the SD had very poor utility. A smaller set of margins might be identified that would be sufficient to model the joint distribution.
 
 The following questions also need to be answered for this method:
\begin{itemize}
\item{How would this method perform for different sample sizes of the GT data? It would be expected to work best with large data sets. But how small would the data sets have to be before coarsening destroys the utility of the SD?}
\item{Could a powerful algorithm reverse-engineer the coarsened margins to recover the counts that lay behind them? A table coarsened to multiples of 10 with $p$ cells would correspond to $10^p$ possible data sets. But the constraints imposed by making the counts add to the GT sample size and to match lower order margins will restrict this number.}
\item{NSAs now can provide a flexible table builder that allows the user to create their own tables. See the National Records of Scotland's plan for release of the 2022 Census\footnote{https://www.scotlandscensus.gov.uk/census-results/flexible-table-builder/ accessed 31/05/2026}. Such tables might perhaps be used to create margins by the methods described here. Could SD created from these systems help users awaiting access to TREs to get started on a research project?}
\end{itemize}

\section{Conclusion}
A simple method of creating SD from margins that have been altered to satisfy SDC rules is proposed. It appears to work satisfactorily on one example, but further work is needed to investigate any disclosure that might arise from the release of such SD and to investigate how it could work on different data sets.

\section{Acknowledgements}
I am grateful to 3 anonymous referees for helpful comments on an earlier draft of this paper.

%
%

\begin{thebibliography}{8}
	
\bibitem{Abowd2018}
	Abowd, J. M. (2018), “The U.S. Census Bureau Adopts Differential Privacy,” in Proceedings of the 24th ACM SIGKDD International Conference on Knowledge Discovery \& Data Mining, 2867.

\bibitem{ADRUK}
ADR UK (2023)  An interim ADR UK position statement on synthetic data. Available from https://www.adruk.org/search/?query=synthetic+data,  accessed 31/05/2026. 

\bibitem{mipfp}
Barthélemy J, Suesse T (2018). “mipfp: An R Package for Multidimensional Array Fitting and Simulating Multivariate Bernoulli Distributions.” Journal of Statistical Software, Code Snippets, 86(2), 1–20. doi:10.18637/jss.v086.c02.

\bibitem{lowfid}
Bharat S.S., Frayling L., Stock J., Lugg\-Widge F., Gordon E., Oliver E.  (2025) A Review of Synthetic Data Terminology for Privacy Preserving Use Cases, Int J Popul Data Sci. \textbf{10:2:08}

\bibitem{synSIPP}
Benedetto G.,  Stanley J.C,, and Totty E. (2018) The Creation and Use of the SIPP Synthetic Beta v7.0, CES Technical Notes Series from Center for Economic Studies, U.S. Census Bureau.

\bibitem{Daniel}
Daniel, O (2025) Is it really private if you can’t explain it? A practical framework for
productionalising legally-compliant synthetic data in government.
UNECE, Expert meeting on Statistical Data Confidentiality, Barcelona. 
Available from https://unece.org/sites/default/files/2025-10/SDC2025\_Sa\_UnitedKingdom\_Daniel.pdf, accessed 26/5/26.

\bibitem{DARE} 
DARE UK (2025) Synthetic data Community Group. Perspectives and Recommendations on the Development of Synthetic Datasets in Trusted Research Environments \url{https://portal.dementiasplatform.uk/reports/development of  synthetic datasets in trusted research environments/}, accessed 27/5/2026.

\bibitem{DARE2} 
DARE UK (2026) Synthetic data Community Group. Synthetic Data Release Framework under UK Data Protection Law  https://zenodo.org/records/20535817, page 8, accessed 29/06/2026.

\bibitem{Dre}
Drechsler, J. (2023). Differential Privacy for Government Agencies—Are We There Yet? Journal of the American Statistical Association, 118(541), 761–773. https://doi.org/10.1080/01621459.2022.2161385

\bibitem{DandS}
Deming, W. E. and Stephan, F. F. (1940). Ann. Math. Statist., \textbf{11, 427–444}.


\bibitem{DandH}
Drechsler J.  Haensch C.A. (2024)
30 Years of Synthetic Data 
Statistical Science \textbf{39,  2}, 221–-242 https://doi.org/10.1214/24\-STS927 , accessed 28/5/2026

\bibitem{ElliotPSD2024}
Elliot, M., Little, C.,  Allmendinger, R. (2024). The Production of Bespoke Synthetic Teaching Datasets Without Access to the Original Data. In M. Önen and J. Domingo-Ferrer (Eds.), Privacy in Statistical Databases, PSD 2024 (Vol. 14915, pp. 144–157). Springer. https://doi.org/10.1007/978\-3\-031-69651\-0\_10


\bibitem{Fienberg}
 Fienberg, S. E. (1970). Ann. Math. Statist., \textbf{41, 907–917.}

\bibitem{FandD}
Fössing, E.,  Drechsler, J. (2024). An Evaluation of Synthetic Data Generators Implemented in the Python Library Synthcity. In M. Onen and J. Domingo-Ferrer (Eds.), Privacy in Statistical Databases, PSD 2024 \textbf{14915,178–193}. Springer. 

\bibitem{res_ready}
Grath-Lone LM, Jay MA, Blackburn R, Gordon E, Zylbersztejn A, Wiljaars L, Gilbert R. (2022) What makes administrative data "research-ready"? A systematic review and thematic analysis of published literature. Int J Popul Data Sci. Apr \textbf{27;7(1):1718}. https://ijpds.org/article/view/1718, accessed 4/3/2025. 10.23889/ijpds.v7i1.1718


\bibitem{frPSD2024}
Green, E., Ritche, F., White, P. (2024). The statbarn: A New Model for Output Statistical Disclosure Control. In M. Önen amd J. Domingo-Ferrer (Eds.), PRIVACY IN STATISTICAL DATABASES, PSD 2024 textbf{14915, 284–-293}. Springer. 

\bibitem{framework}
Houssiau, F., Cohen, S. N., Szpruch, L., Daniel, O., Lawrence, M. G., Mitra, R., Wilde, H.,  Mole, C. (2022). A Framework for Auditable Synthetic Data Generation. https://doi.org/10.48550/arxiv.2211.11540, Accessed June 2026,

 \bibitem{synLBD}
Kinney, S.K., Reiter, J.P., Reznek, A. P., Miranda, J., Jarmin, R., Abowd, J.M. (2011), Towards Unrestricted Public use Business Microdata: The Synthetic Longitudinal Business Database, International Statistical Review, \textbf{79 (3), 362-384}.

\bibitem{ashe_census2011}
Little C., Elliot M., Allmendinger, M. (2024)
USER GUIDE: Synthetic ASHE-2011 Census dataset 
 DOI: http://doi.org/10.5255/UKDA-SN-9282-1
 
 \bibitem{LittleWu}
Little, R. J. A.,  Wu, M.-M. (1991). “Models for Contingency Tables With Known Margins When Target and Sampled Populations Differ”: TM. Journal of the American Statistical Association, 86(413), 87.
 

\bibitem{McKenna}
McKenna, R., Miklau, G. and Sheldon, D. (2021) “Winning the NIST Contest: A scalable and general approach to differentially private synthetic data”, Journal of Privacy and Confidentiality, 11(3). doi: 10.29012/jpc.778.

\bibitem{Nowok}
Nowok B, Raab GM, Dibben C (2016). “synthpop: Bespoke Creation of Synthetic Data in R.” Journal of Statistical Software, 74(11), 1–26. doi:10.18637/jss.v074.i11. The ipf method was added in Version 1.50  in 2018.

\bibitem{ONS}
Office of National Statistics (2023), Synthesising the linked 2011 Census and deaths dataset while preserving its confidentiality, ONS Data Science Campus, United Kingdom. https://datasciencecampus.ons.gov.uk/synthesising-the-linked-2011-census-and-deaths-dataset-while-preserving-its-confidentiality/ accessed 29/06/2026.

\bibitem{owen}
Owen, D. (2025) Is it really private if you can’t explain it? A practical framework for productionalising legally-compliant synthetic data in government.
UNECE Expert Meeting on Statistical Data Confidentiality 15-17 October 2025, Barcelona. 

\bibitem{SDV}
Patki N,  Wedge R, Veeramachaneni K, "The Synthetic Data Vault," 2016 IEEE International Conference on Data Science and Advanced Analytics (DSAA), Montreal, QC, Canada, 2016, pp. 399-410, doi: 10.1109/DSAA.2016.49.

\bibitem{RaabDP}
Raab, G. M. (2022). Utility and Disclosure Risk for Differentially Private Synthetic Categorical Data. In J. Domingo-Ferrer and M. Laurent (Eds.), Privacy in Statistical Databases (Vol. 13463, pp. 250–265). Springer International Publishing AG. 

\bibitem{RaabUNECE2025}
Raab G.M., Dibben C., Krčo N. (2025) Confidentiality and disclosure risk from administrative data 
UNECE, Expert meeting on Statistical Data Confidentiality, Barcelona. 
Available from 
https://unece.org/sites/default/files/2025\-10/SDC2025\_Sb\_UnivEd-SLS\_RaabDibbenKcro\_D.pdf, accessed 26/5/26.

\bibitem{Fourchecks}
Raab, G., McCall, S. and Cavin, L. (2025) “Four checks for low\-fidelity synthetic data: recommendations for disclosure control and quality evaluation”, International Journal of Population Data Science, 10(2). doi: 10.23889/ijpds.v10i2.2972.

\bibitem{Raab2021}
Raab G.M., Nowok B and Dibben C. (2021)
Assessing, visualizing and improving the utility of synthetic data, preprint available from https://arxiv.org/abs/2109.12717, accessed 06/2026.


\bibitem{RaabPSD2024}{
Raab, G. M. (2024). Privacy Risk from Synthetic Data: Practical Proposals. In M. Onen and J. Domingo-Ferrer (Eds.), Privacy in Statistical Databases PSD2024 \textbf{14915, 254--273}. Springer. 
}
\bibitem{Ritchie2007}
Ritchie F. (2007) Disclosure detection in research environments in practice. Paper presented at UNECE/Eurostat work session on statistical data confidentiality - 2007.



\bibitem{Rubin}
Rubin, D. (1993) Discussion: Statistical Disclosure Limitation.  Journal of Official Statistics. \textbf{9} 461–-468.

\bibitem{Rug}
Ruggles, S., Fitch, C., Magnuson, D., and Schroeder, J. (2019), “Differential Privacy and Census Data: Implications for Social and Economic Research,” in AEA Papers and Proceedings (Vol. 109), pp. 403–408.

\bibitem{smithsdc}
Smith, J., Padiya, T., Ritchie, F., Green, E.,  Tilbrook, A. (2025) A formal model for reasoning about output disclosure risks and mitigations. UNECE Expert meeting on Statistical Data Confidentiality, Barcelona.  Available from https://uwe\-repository.worktribe.com/output/15152512, accessed 26/5/26.


\bibitem{snoke}
Snoke J, Raab G, Nowok B, Dibben C, Slavkovic A (2018). “General and Specific Utility
Measures for Synthetic Data.” Journal of the Royal Statistical Society B, textbf{181(3)}, 663-–668.

\bibitem{Stephan}
Stephan, F. F. (1942). An Iterative Method of Adjusting Sample Frequency Tables When Expected Marginal Totals are Known. The Annals of Mathematical Statistics, 13(2), 166–178.

\bibitem{Templ}
Templ, M., Meindl, B., Kowarik, A., \& Dupriez, O. (2017). Simulation of Synthetic Complex Data: The R Package simPop. Journal of Statistical Software, 79(10), 1–38. https://doi.org/10.18637/jss.v079.i10

\bibitem{Newton}
Taub J, Elliot M, Raab GM, Chareset A, Chen C, O’Keefe CM, Pistner M, Snoke J, Slavkovic A (2019) Creating the Best Risk-Utility Profile: The Synthetic Data Challenge, Joint UNECE/Eurostat Work Session on Statistical Data Confidentiality. 

\bibitem{VandW}
Voas D, Williamson P (2001). “Evaluating Goodness-of-Fit Measures for Synthetic Microdata.” Geographical and Environmental Modelling, \textbf{5}, 177-–200.


\bibitem{Wink}
Winkler, R. L., Butler, J. L., Curtis, K. J., and Egan-Robertson, D. (2021), “Differential Privacy and the Accuracy of County-Level Net Migration Estimates,” Population Research and Policy Review, 41, 417–435.


\bibitem{NZ}
Young J. Graham P. Penny, R. (2009). Using Bayesian Networks to Create Synthetic Data. Journal of Official Statistics. \textbf{25. 549--567}. 




\end{thebibliography}
%

\end{document}